\documentclass[runningheads]{llncs}
\usepackage{graphicx}
\usepackage{amsmath,amssymb} 
\usepackage{color}
\usepackage{algorithm}
\usepackage{algpseudocode}

\algdef{SE}[DOWHILE]{Do}{doWhile}{\algorithmicdo}[1]{\algorithmicwhile\ #1}%

\usepackage{sidecap}
\usepackage{wrapfig}
\usepackage{lipsum}

\begin{document}
\pagestyle{headings}
\mainmatter


\title{Who's that Actor? Automatic Labelling of Actors in TV series starting from IMDB Images} 
\authorrunning{Rahaf Aljundi, Punarjay Chakravarty and Tinne Tuytelaars}
\titlerunning{Who's That Actor? Automatic Labelling of Actors in TV series.}

\author{Rahaf Aljundi\thanks{Authors with equal contribution}, Punarjay Chakravarty$^{ 	\star}$  and Tinne Tuytelaars}
\institute{KU Leuven, ESAT-PSI, iMinds, Belgium}

\maketitle

\begin{abstract}
In this work, we aim at automatically labeling actors in a TV series.
Rather than relying on transcripts and subtitles, as has been demonstrated in the past,
we show how to achieve this goal starting from a set of example images 
of each of the main actors involved, collected from the Internet Movie Database (IMDB).
The problem then becomes one of domain adaptation:
actors' IMDB photos are typically taken at awards ceremonies and are quite different
from their appearances in TV series. In each series as well, there is considerable
change in actor appearance due to makeup, lighting, ageing, etc. 
To bridge this gap, we propose a graph-matching based self-labelling algorithm,
which we coin HSL (Hungarian Self Labeling). Further, we propose a new edge cost 
to be used in this context, as well as an extension that is more robust to outliers, where prototypical
faces for each of the actors are selected based on a hierarchical clustering procedure.
We conduct experiments with 15 episodes from 3 different TV series and 
demonstrate automatic annotation with an accuracy of 90\% and up. 

%
\end{abstract}

\section{Introduction}
There has been an explosion of video data in the recent past. In addition to  professionally shot movies and TV series, with the proliferation of smart phones and the advent of social media, users document more and more of their lives on home-video. To properly use this data, like the search engines of the world wide web, there is a need
for archiving and indexing these videos for easy search and retrieval.
To this end, we present an automatic person labelling system in video starting from only a few sample images, which
could be actor images from IMDB (as in our experiments) or ``tagged'' images on social media.
Our system can be used to archive, index and search large databases of video using just a few images of the main characters as starting point. Using our method, one can imagine a video search app that could for example search for the scene in the TV series Breaking Bad, where Walter White first meets Jesse Pinkman, or for all videos of grandpa, from a collection of home videos. \newline
The first major attempt at automatically labelling actors in a TV series was made by Everingham et al.~\cite{everingham2006hello}, who in their seminal work demonstrated
the training of classifiers for actor recognition using the weak supervision of subtitle and transcript files.  
However, subtitles and transcripts are not always available. They can be hard to find for some movies and TV shows, and are perhaps non-existent for TV news broadcasts, music-videos, documentaries and silent films.
Even if they are available, transcripts, in most cases obtained from fan websites, vary in quality and reliability. They also come in different formats, making it hard to automate the process. 
In addition,
there is the need to align the transcripts with the subtitles. 
In the absence of transcripts, \cite{tapaswi2012knock,hu2015deep,ren2016look} use supervision from hand-labelled ground-truth for some of the video data. Collecting such annotations is, however, a cumbersome and user-unfriendly process.\newline
In this paper, we consider an alternative source for training appearance models for actors, the Internet Movie Database, or IMDB. Most, if not all TV series are listed on IMDB, along with photos of the major actors. Our method is simple to use. All it needs is the name of the show. If the title exists on IMDB, the downloading of actor images and their use in the labelling of actor appearances in the TV show or movie is completely automatic.\newline
\begin{figure}[t!]
 \vspace*{-15pt} 
\begin{center}
    \includegraphics[width=0.99\linewidth]{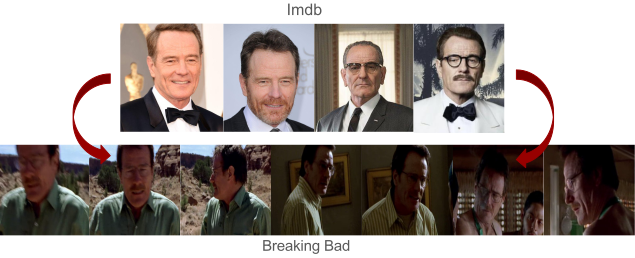} 
\end{center}
 \vspace*{-15pt} 
   \caption{The problem of actor labelling using IMDB images is one of accounting for the domain shift between photos in IMDB and frames from the TV show.}
\label{fig:domainShift}
 \vspace*{-15pt} 
\end{figure}
This may seem straightforward, but actually it is not.
The images in IMDB are mostly taken at awards functions and ceremonies using flash-enabled DSLR cameras, or are from promotional material for the shows. An actor's appearance in a show is normally quite different from her appearance in IMDB. This is due to various factors like camera motion (and consequent blur), lighting changes, make-up of the actor, and in-show or real-life ageing of the star. In other words, there is a {\em domain shift} between the IMDB data and the data from the TV shows. In this work, we aim at overcoming this domain shift (see Figure ~\ref{fig:domainShift}). \newline
In a self-labelling inspired approach, we iteratively select actors' faces from the TV series that are closest to their counterparts in IMDB and use them to enrich the set of IMDB template images. As more and more actor faces from the show are added to the original set of IMDB templates, the actor's appearance in the show is better represented, the domain shift is accounted for, and the method is better able to label the actor inspite of major changes in appearance compared to the original IMDB templates.\newline
To optimally select the set of images in each step of the self-labelling procedure, we propose to use bipartite graph matching and incorporate a new graph edge cost. In order to strengthen the method's stability in various challenging cases (pose change, lighting change, presence of side actors, etc.), we further add a clustering/outlier removal scheme resulting in an {\em actor profile} that is used for face labelling in later iterations. 
We utilize the power of the latest generation face recognition and verification techniques - Deep Face features \cite{Taigman2014deepface,Parkhi15,schroff2015facenet}. These features are activations of Convolutional Neural Networks trained on vast face recognition databases. They are able to characterize a person under a wide range of poses and lighting conditions. We use a multi-target tracker to simultaneously track multiple people in video. Each track has many faces of the same person under different poses and our labelling algorithm takes advantage of the information available from all the faces in a track to identify it. We do not use any context information such as clothing.\newline
We demonstrate actor labelling in the TV series Big Bang Theory, Breaking Bad and Mad Men. For each of these
series, we manually download a small number of template images of the top 5/6 actors from IMDB and demonstrate an average of 91\% accuracy in detecting those characters in these TV series.\newline
In summary, our main contributions are : 1) a new graph edge cost that improves the selection of hitherto unlabelled faces in the self-labelling process; 2) the use of bipartite graph matching for optimal selection of unlabelled faces in each step of the self-labelling process; 3) a hierarchical clustering procedure yielding an actor profile and increasing the robustness of the method against outliers and side-actors; and 4) a dataset comprising of labelled faces in a total of 15 episodes from 3 TV series.\newline
We detail prior related work in Section \ref{RelatedWork},  describe our system in Section \ref{SystemDescription}, report experiments in Section \ref{Experiments}, discuss results in Section \ref{ResultsAndDiscussion} and conclude with Section \ref{Conclusion}.
\section{Related Work}
\label{RelatedWork}
\textbf{Weak Supervision for Actor Labelling} 
The topic of labeling actors in movies and TV-series has been tackled mostly using weak supervision from subtitle and transcript files~\cite{everingham2006hello,sivic2009you,bauml2013semi,tapaswi2015improved,bojanowski2013finding,Parkhi15a,hauriletnaming}, building on the seminal work of
Everingham et al.~\cite{everingham2006hello,sivic2009you}.
The transcript tells us what each character says in the show, but is not aligned with the video. On the other hand, the subtitle file is aligned with the video, but does not have speaker labels. \cite{everingham2006hello,sivic2009you} proposes to combine the two (using words in the dialogues) to get the label for an active speaker, if one exists, in each frame. Active speakers detected by lip movement detection in video can then be labelled from the aligned subtitle and transcript files. This weak supervision allows training face and clothing based classifiers for each person based on which they can be labelled even in the absence of speech.

Tapaswi et al.~\cite{tapaswi2015improved} build on this work, using additional cues to train classifiers in a joint optimization framework, such as the fact that no two people in the same scene can have the same label.
Bauml et al.~\cite{bauml2013semi} employ an additional class for side-actors and take into account unlabelled data (faces without transcript associations) by using an entropy function that encourages classifier decision boundaries to lie along low-density areas of unlabelled data points. 
Both~\cite{bauml2013semi} and~\cite{tapaswi2015improved} report average accuracy results of around 83\% on the first 6 episodes of the TV series The Big Bang Theory.


Building on the work of~\cite{guillaumin2010multiple}, the assignment of actor names in transcripts to one (of several) people in the frame has also been treated as a Multiple Instance Learning (MIL) problem~\cite{Parkhi15a,hauriletnaming}. 
They both use CNN features as face descriptors.
\cite{Parkhi15a} also shows that having a separate side-actor classifier boosts actor recognition performance.
Finally, action recognition has been used as an additional cue as well, along with weak supervision from transcripts~\cite{bojanowski2013finding}. 

In contrast to this previous work on actor labelling, our supervision is not from transcripts and subtitle files, which can be difficult to obtain and align with video, but from IMDB actor images. The names of the main actors are used to obtain a few IMDB images for weak supervision, and then propagated through the video data via self-labelling.

\textbf{Label Propagation in Domain Adaptation}
Label propagation is widely used in semi-supervised learning, i.e. dealing with the case when there is a mix of labelled and unlabelled data. In one of the first papers on label propagation, \cite{zhu2002learning} builds a graph with weighted edges indicating the similarity between nodes. Some of the nodes are labelled, and these labels are propagated to their neighbouring, unlabelled nodes according to some similarity measure. The labels propagate through the graph while preventing the initially given labels from changing.

Pham et al.~\cite{tuytelaars2011naming} use label propagation to name people appearing in TV news broadcasts starting from transcripts for weak supervision. Kumar et al.~\cite{kumar2014face} also use label propagation to propagate actor labels from a set of key frames that are matched to a manually-curated selection of template images in a movie. However, in their case, and in semi-supervised learning in general, it is supposed that the source and target data are from the same distribution. This is not the case in domain adaptation (DA, see \cite{patel2015visual} for a survey)  where there is a domain shift between the labeled source data and unlabelled target data.
We use self-labelling~\cite{bruzzone2010domain}, which is a variant of label-propagation in the presence of domain shift, to deal with this. In our case, the domain shift is between the IMDB images that are used for supervision, and the actor faces in the TV series.
Bruzzone et al.~\cite{bruzzone2010domain} suggest an iterative labelling strategy that uses the initial labelled source data to train an SVM model and then gradually add the target data obtained from self-labelling to adapt the decision function (i.e the learned model), while simultaneously slowly removing the source labels. The final classifier is only learnt from the self-labelled target samples. A theoretical guarantee for the self-labelling DA to work has been provided in~\cite{habrard2013iterative}. 

We incorporate bipartite graph matching to optimally select the best unlabelled faces to be added to the actor template images in each iteration of the self-labelling process. Graph matching has been used for DA before in remote sensing~\cite{banerjee2015novel,tuia2013graph}. For example, in~\cite{tuia2013graph} the authors match the built graphs in the source and target domains after non linearly transforming (aligning) the source domain with the target domain. However, to the best of our knowledge, the use of graph matching within self-labelling has not been studied before.

\section{System Description}
\label{SystemDescription}
We first describe the data preprocessing: face detection, description and tracking (Section~\ref{sec:preprocessing}). Then, we move on to the graph-matching based self-labelling, including the new graph edge cost and the hierarchical clustering extension (Section~\ref{sec:labelling}).

\subsection{Preprocessing}
\label{sec:preprocessing}
\noindent
{\bf Face Detection and Face Feature Extraction}
We use a Deformable Parts Model (DPM)~\cite{mathias2014face,voc-release5} for face detection and a pretrained CNN model (Deep Face VGG model~\cite{Parkhi15}) for face description. The latter model has been trained for face recognition on a database of 2622 celebrities (with 375 face images each). We use the output of the last but one fully connected layer (after L2-normalization), which gives us a 4096 dimensional descriptor, used both for tracking faces and for matching actor face tracks to faces in the IMDB database. \\

\begin{figure}[t]
\begin{center}
    \includegraphics[width=1\linewidth]{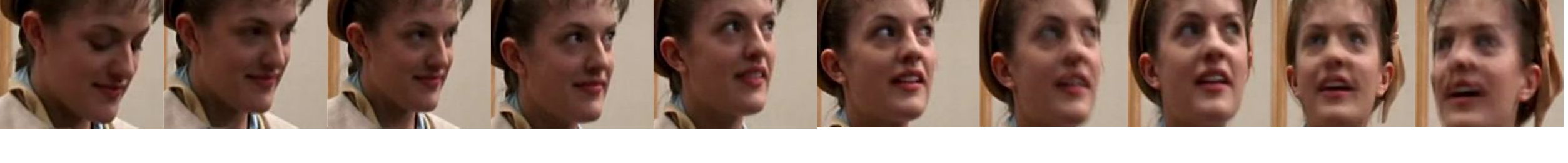} 
\end{center}
 \vspace*{-0.6cm} 
   \caption{Different faces and poses in a single track
   (Breaking Bad, Episode 1).
   }
\label{fig:tracking}
 \vspace*{-0.6cm} 
\end{figure}

\noindent
{\bf Tracking}
We employ a multi-target tracker for tracking faces in the TV series. The tracker receives as  input the face detections in each frame, along with their CNN feature vectors. New tracks are initialized at detections that do not overlap with previously operating tracks. Existing tracks are updated as follows:
if the bounding box of the track and a detection are overlapping, and the  Euclidean distance between their feature vectors is below a threshold, the track's coordinates are updated to those of the bounding box of the detection.
When there are multiple tracks and detections in close proximity, the Hungarian algorithm~\cite{kuhn1955hungarian} is used to optimally associate tracks and detections based on the matching scores (Euclidean distances) of their feature vectors, avoiding multiple detections being assigned to the same track, or a single detection being assigned to multiple tracks. Only associations that are below the previously mentioned threshold distance are considered  in the Hungarian optimization.

Tracks that are not updated for a threshold number of frames are deleted. Tracks are also deleted at shot boundaries. Shot boundary detection is done by a simple histogram comparison between frames.

Tracking gives us a sequence of faces in time, that are similar in face descriptor space (see Figure~\ref{fig:tracking}). We adjust our Euclidean distance threshold (for similarity between face descriptors) so that we err on the side of more tracks of the same actor in the same shot (early track termination) instead of merging tracks of different actors.
The advantage with tracking is that the labelling algorithm has the information from all the faces  in the track to make a decision about the track identity. 

~\subsection{H(C)SL: Hungarian (Clustering-based) Self-Labelling}
\label{sec:labelling}
Tracking the detected faces gives us a set of tracks for a given video/episode. These
tracks belong to the main actors, whose images we have from IMDB, as well as side actors, for whom we do not have profile images. Each track is composed of a set of detected faces and we consider each track as one entity.  
As explained earlier, we also have from IMDB, a set of faces for each given main actor. We consider the set of faces of each actor from IMDB as a point cloud in feature space. The appearances of these faces are different from the actors' appearances in the tracks, because of reasons explained earlier. 
Because of this domain shift,
a naive nearest neighbour matching of the tracks to the closest IMDB faces,
will not give good results.
We later illustrate this with a baseline experiment.

In spite of the variation between the tracks and the IMDB faces, there still exist some tracks
that look sufficiently similar to the IMDB faces to be labelled correctly.
To successfully label all the tracks, we make use of these best matching tracks, gradually adding tracks to the actor cloud in a strategy inspired by self-labelling.\\

\noindent
{\bf Graph-based matching using the Hungarian algorithm}
We now explain our method for selecting tracks and moving them to the actor cloud.
Note that, in each iteration, at most one track is added to each actor's cloud (face set). 

We first construct a bipartite graph  $G=(V,E)$  where vertices represent sets of faces (actor clouds or tracks), and edges indicate an assignment (at a certain cost) of a track to an actor cloud. $V$ consists of two sets: the actors set, $V_{Actors}=\{V^i_{Actors}\}$, and the set of tracks, $V_{Tracks}=\{V^j_{Tracks}\}$. The number of vertices in $V_{Actors}$ corresponds to the number of main actors. The number of vertices in $V_{Tracks}$ is much larger. Each IMDB actor vertex represents the set of faces that correspond to that actor:
\begin{equation}
V_{Actors}^i=\{ f^i_{1},\cdots,f^i_{im},\cdots, f^i_{n_{im}} \}  , \textrm{with } n_{im} \textrm{ the number of faces  of  actor } i. 
\end{equation} 
Each vertex in $V_{tracks}$ represents a track extracted from the TV series:
\begin{equation}
V_{Tracks}^j=\{ f^j_{1},\cdots, f^j_{tr},\cdots, f^j_{n_{tr}} \}  , \textrm{with } n_{tr} \textrm{ the number of faces  in  track } j. 
\end{equation}

\begin{figure}[t]
\centering
\includegraphics[width=0.9\textwidth,scale=0.7]{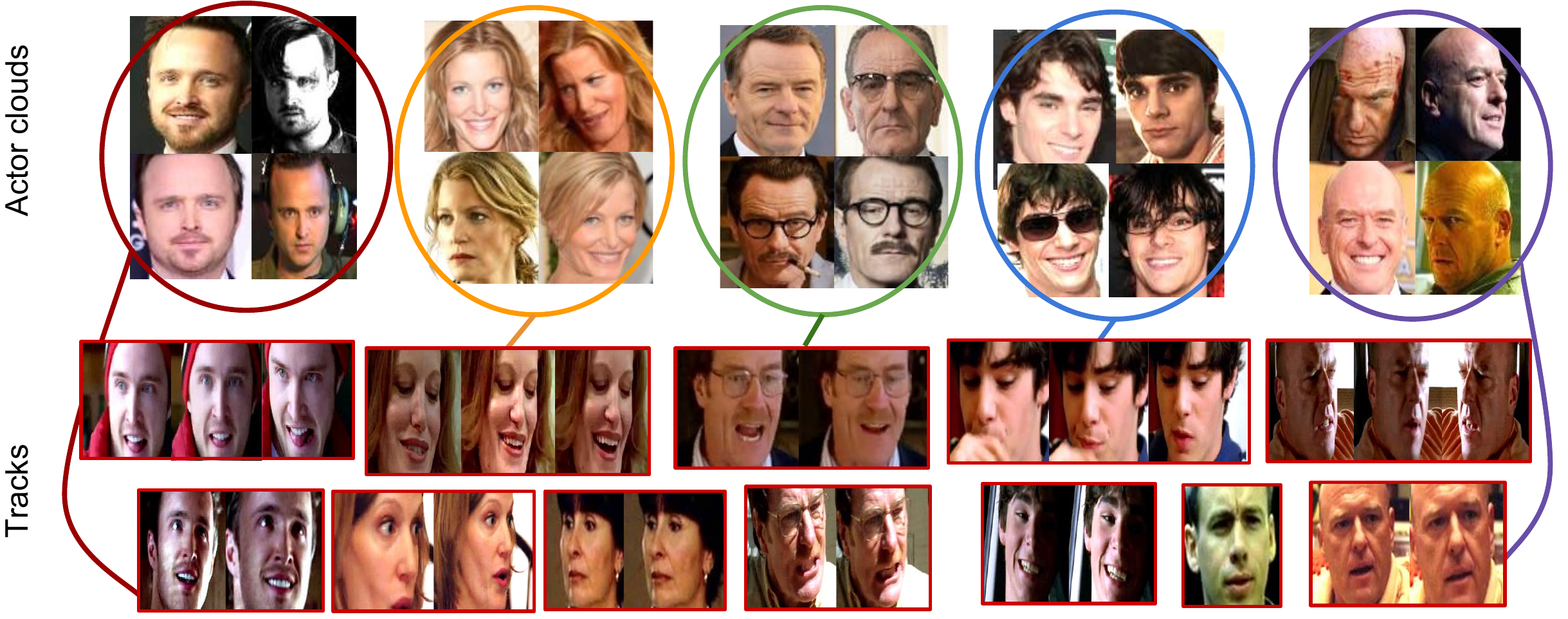}
\caption{Hungarian graph matching between vertices in the actors set (top row) and tracks from the video (bottom rows). Matched tracks at this iteration are indicated with solid connections.}
 \vspace*{-20pt}
\label{fig:hung1}
\end{figure}

Each vertex in $V_{Tracks}$ can have an assignment (i.e. an edge) to one and only one vertex in $V_{Actors}$. It is possible that a $V_{Tracks}$ vertex remains unassigned, for example if the track belongs to a side actor. 
While in theory a  $V_{Actors}$ vertex can be linked to more than one vertex in $V_{Tracks}$,
we only allow one edge to it at each step in our implementation.  
Each edge  describes the cost of matching the track to the actor. Our task is to find the  optimal matching of the tracks to the actors. To do so, we use the Hungarian method~\cite{kuhn1955hungarian} that assigns to each actor $V_{Actors}^i$, a track $V_{Tracks}^j$, as illustrated in Figure~\ref{fig:hung1}. 
We set up a cost matrix with each element in the matrix representing the matching cost between an actor and a track, $w(V^j_{Tracks},V^i_{Actors})$. This matching cost $w$ is given by equation \ref{eq:normalizedCost2} below.



\noindent 
{\bf Graph Edge Cost}
The choice of the cost measure $w(V^j_{Tracks},V^i_{Actors})$ in the Hungarian cost matrix, with $V^j_{Tracks}$ and $V^i_{Actors}$ sets of faces, is crucial as it is the main ingredient of the selection process. Rather than using the minimum or average distance as a cost, we introduce the {\em normalized edge cost} between a track and a set of faces. To this end, we first compute the average distance $d$ between the faces in the track and the actor faces in the actor cloud:
\begin{equation}
d(V^j_{Tracks},V^i_{Actors})=\frac{1}{n_{tr}} \sum_{f^j_{tr}\in V^j_{Tracks}} \frac{1}{n_{im}}\sum_{f^i_{im}\in V^i_{Actors} } \parallel f^j_{tr}-f^i_{im}\parallel^2
\label{eq:normalizedCost1}
\end{equation}
Then the cost of assigning a track $j$ to an actor $i$ is this average distance, followed by the subtraction of the mean of the average distance of the track $j$ to all actors:
\begin{equation}
w(V^j_{Tracks},V^i_{Actors})=d(V^j_{Tracks},V^i_{Actors}) -\frac{1}{n_{ac}}\sum _i d(V^j_{Tracks},V^i_{Actors})
\label{eq:normalizedCost2}
\end{equation}
where $n_{ac}$ is the  number of main actors. This encourages the selection of face tracks that are closer to a specific  actor cloud compared to others, while tracks that are equidistant from all actor clouds, typically side-actors or 
faces recorded under atypical conditions,
are less likely to be selected. \\


\noindent
{\bf Hungarian Self-Labelling}
Having explained the graph matching step,  let us summarize the steps of the proposed method. We start with actor clouds containing only the  IMDB faces. Then, we iteratively assign tracks from the TV series to their corresponding actor clouds using Hungarian graph matching (at most one track per actor at a time), and move the faces in these tracks to the corresponding actor clouds. We repeat this process until 
no more tracks can be assigned.
The iterative assignment step reduces the distance between the actor clouds and the remaining TV series tracks and gradually removes the domain shift. As a result, with each successive iteration of graph-matching based self-labelling, the process is able to label tracks in the TV series that contain faces that are more and more different from the original IMDB images.
If there is a mismatch in the number of tracks between actor A and actor B - say 500 tracks for actor A and 100 tracks for actor B - after all the tracks for actor B have been labelled, the Hungarian method will still need to assign tracks for actor B. These will be tracks that don't actually belong to actor B. To prevent this, we use a threshold ($\lambda$) for track assignment. 
Algorithm \ref{algo1}  shows the steps of the proposed graph-matching based self-labelling procedure.
\begin{algorithm}

  \caption{Hungarian self-labelling ($V_{Tracks},V_{Actors},\lambda$)}\label{algo1}
\begin{algorithmic}[1]
  \State $V_{Tracks}={V^1_{Tracks},V^2_{Tracks},..,V^m_{Tracks}}$\Comment{ the set of tracks, m: the number of tracks}
  \State $V_{Actors}={V^1_{Actors},V^2_{Actors},..,V^n_{Actors}}$\Comment{ the actor clouds, n: the number of actors}
 \Do
      \State $number\_ of\_ added\_ tracks = 0$
      \State $w$=compute\_cost($V_{Tracks}$,$V_{Actors}$) \Comment {the cost of assigning each track to each
      \State    $\,\,\,\,\,\,\,\,\,\,\,\,\,\,\,\,\,\,\,\,\,\,\,\,\,\,$  $\,\,\,\,\,\,\,\,\,\,\,\,\,\,\,\,\,\,\,\,\,\,\,\,\,\,$  $\,\,\,\,\,\,\,\,\,\,\,\,\,\,\,\,\,\,\,\,\,\,\,\,\,\,$  $\,\,\,\,\,\,\,\,\,\,\,\,\,\,\,\,\,$ actor (equation \ref{eq:normalizedCost2})}
      \State assignments=hungarian($V_{Tracks}$,$V_{Actors}$,w)
      \For {$V^i_{Actors}$ in $V_{Actors}$ }
      \State best\_track=assignments($V^i_{Actors}$)
      \If {$w(V^{best\_track}_{Tracks},V^i_{Actors})< \lambda$}
      \State $V^i_{Actors}=(V^i_{Actors} \cup V^{best\_track}_{Tracks}$)  
      \State  $V_{Tracks}=V_{Tracks} \setminus \{V^{best\_track}_{Tracks}\}$ 
      \State $number\_ of\_ added\_ tracks = number\_ of\_ added\_ tracks + 1$
      \EndIf
      \EndFor
    \doWhile{$number\_ of\_ added\_ tracks>0$}
\end{algorithmic}
\end{algorithm}

It is worth noting that the domain shift between IMDB and TV series is not constant, even for the same actor. Hence, it is difficult to come up with a global transformation to overcome this shift. We performed some experiments with Transfer Joint Matching \cite{Long_2014_CVPR}, which is a global optimization procedure, but it took 60x longer to run, and gave inferior results to our baselines.





\begin{wrapfigure}{r}{0.5\textwidth}
  \vspace{-25pt}
  \begin{center}
    \includegraphics[width=0.48\textwidth]{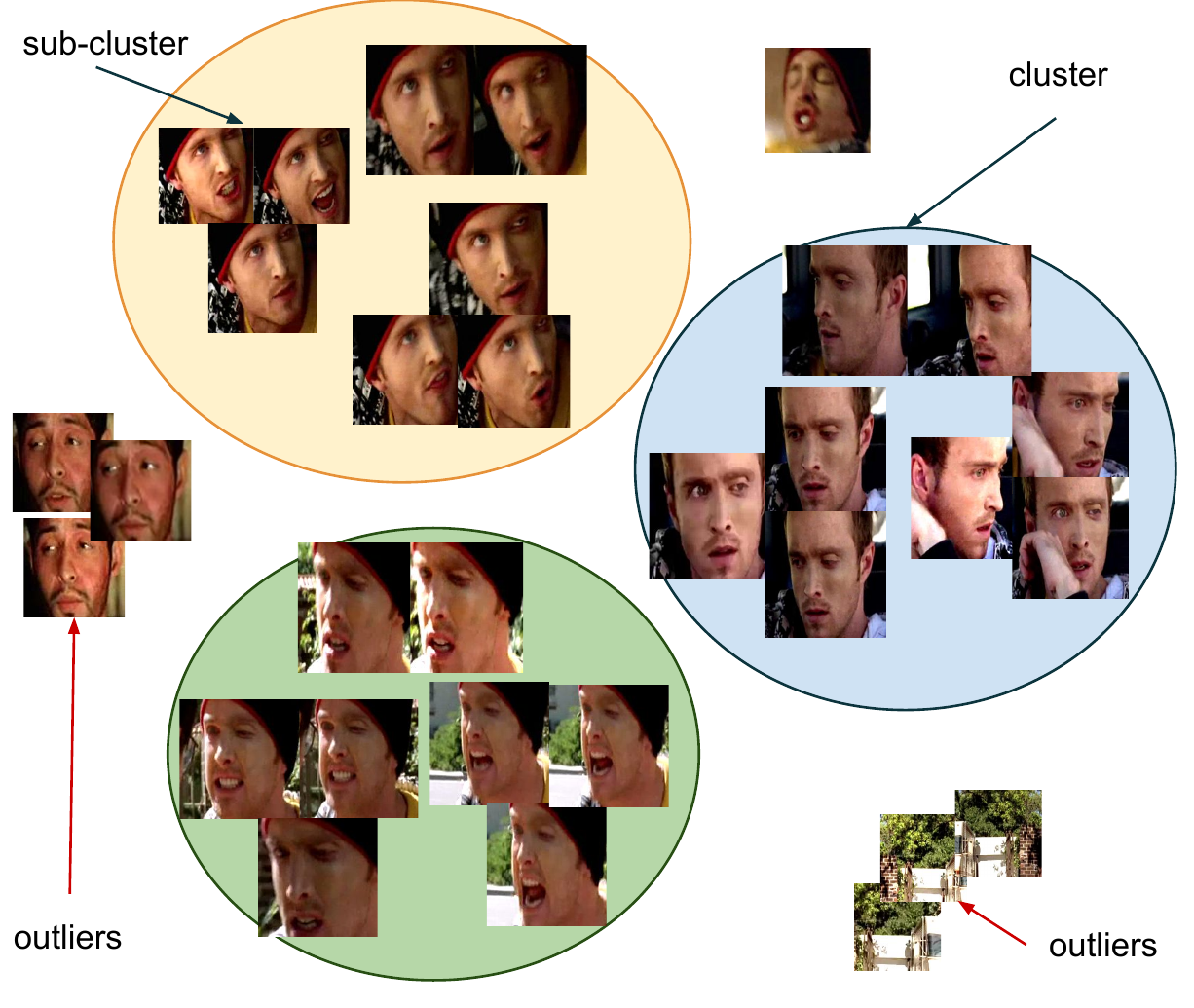}
  \end{center}
  \vspace{-15pt}
  \caption{Clustering of track faces into a cloud - comprising of clusters, sub-clusters and outliers for each actor.}
   \vspace{-25pt}
 \label{fig:clustering}
  
\end{wrapfigure}

\noindent
\textbf{Hierarchical clustering/Outlier removal}
In the self-labelling process, tracks are matched to actors and all the faces in the track are moved to the corresponding actor cloud. However, some of these faces might be outliers: belonging to a false face detection, quite different from most other faces in the cloud, or resembling rare cases of the actor appearance that are extremely different, such as a face in a dimly-lit night scene, blurred or half-occluded.
Including all the faces in the actor cloud (incl. those outliers) in the cost computation for self-labelling might result in false labelling of the remaining tracks. This motivates having a better representation for each actor rather than using all the faces in the actor cloud.
To this end, we use a hierarchical clustering approach that  
\textbf{1.} removes  the outlier faces, and \textbf{2.} selects representative sample points from all the faces representing the actor appearance to obtain a complete profile of the actor, to be used for the cost computation in the self-labelling.
We achieve this in two steps: first we obtain big clusters using a relaxed clustering criterion that excludes all the outlying faces; subsequently, we use a more strict clustering criterion to obtain tighter sub-clusters within each big cluster. The centroids of these sub-clusters are chosen as our set of representative points/faces for the actor under consideration.
Since we do not care about the exact boundaries between the clusters, and do not have to make a decision using points on these boundaries (we only use the centroids), a simple, yet efficient nearest neighbour clustering method \cite{ward1963hierarchical,bubeck2009nearest} is sufficient for our problem.
This hierarchical nearest neighbour clustering is done in an online fashion, during the addition of each track from the TV series to the actor cloud. Figure \ref{fig:clustering} illustrates our clustering procedure.\\

\begin{figure}
 \vspace*{-15pt}
\centering
\includegraphics[width=0.9\textwidth,scale=0.1]{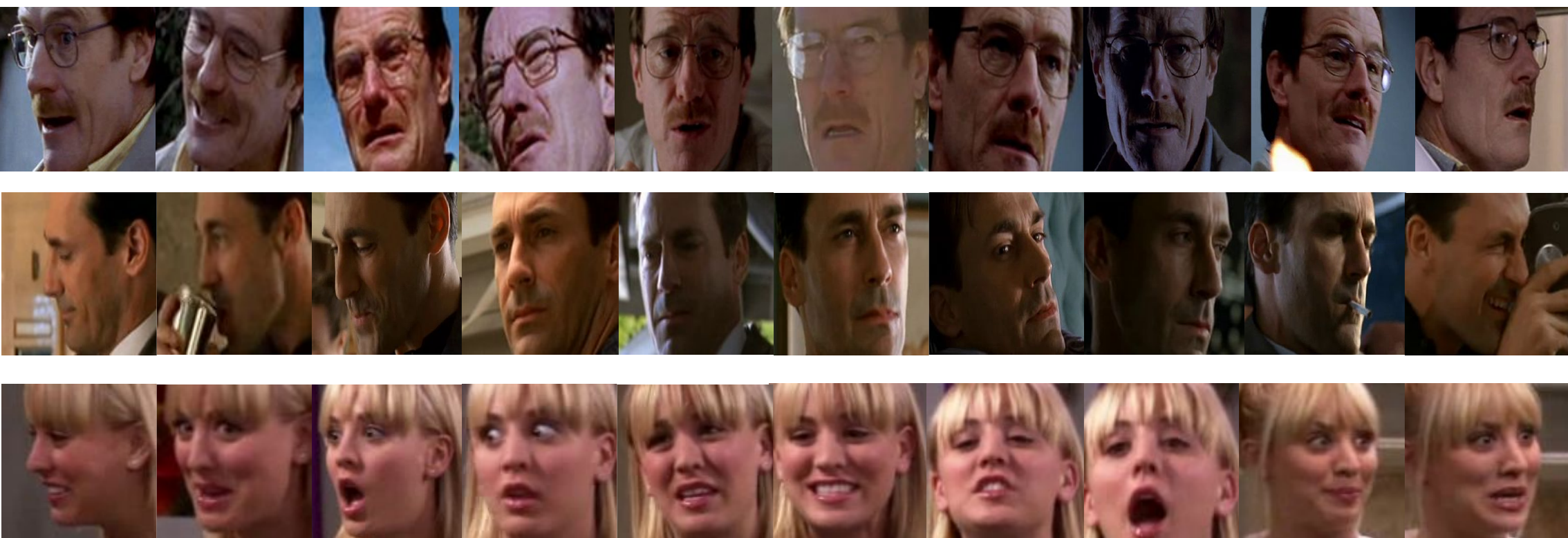}
\caption{From top to bottom: actor profiles for actor Bryan Canston in  Breaking Bad, actor Jon Hamm in  Mad Men and actress Kaley Cuoco in Big Bang Theory,  obtained from the centroids of the sub-clusters.}
\label{fig:actor_profile}
 \vspace*{-10pt}
\end{figure}

%
\noindent
\textbf{Online nearest neighbor clustering}
In our clustering method we leave the original IMDB images out of the calculation: they remain in the cost computation function as regularizers.
So we start with empty clouds for each actor, and receive one track in each assignment step. Each track $V_{Tracks}^j$ is composed of multiple faces$\{f_1^j,f_2^j, \dots, f_n^j\}$, and we process each face independently. The first face $f_1^j$ initializes the first cluster $cluster_1$. The second face $f_2^j$  is then compared with the first cluster $cluster_1$ 
and if the distance is less than a threshold, it is merged with the first cluster. Otherwise, it is assigned to a new cluster.
The same procedure is repeated, until all tracks have been clustered. Small clusters (with number of elements below a threshold) are considered as outliers. For all other clusters, a similar clustering scheme is applied within the clusters, to obtain sub-clusters. The centroids of these sub-clusters are the representative points/faces for each actor cloud.
Figure~\ref{fig:actor_profile} shows some actor profile examples. 

\section{Experiments}
\label{Experiments}
\textbf{Data Set Creation}
We conduct a variety of experiments on three different TV series: Breaking Bad, Big Bang Theory and Mad Men. Each of these series exhibit different characteristics and challenges, from the low quality indoor scenes of Big Bang Theory to the crime drama in Breaking Bad with varied actors looks, light, etc. \textbf{Breaking Bad (BB)} is a crime drama with challenging indoor and outdoor scenes, shaky camera work and a reasonable number of side-actors. We process episodes 1-6 from Season 1. \textbf{Big Bang Theory (BBT)} is a sitcom, shot indoors, with a relatively constant cast, and a smaller number of side-actors. We process episodes 1-6 from Season 1. \textbf{Mad Men (MM)} is a period drama, shot both indoor and outdoors with a large number of side-actors. We process episodes 1-3 from Season 1. Figure \ref{fig:segments}(b) shows the percentage of the tracks that belong to each main actor and those belonging to the side actors in the three TV series.

We put all episodes through the following pipeline:
\begin{enumerate}
\item Face detection using the DPM model  
\item CNN feature extraction 
\item Tracking
\item Repeat 1 and 2 for IMDB face images
\item Label the tracks using IMDB images as supervision
\end{enumerate}

We download a small number (15-20) of IMDB images as templates for each actor and ensure that no images from the TV series themselves are included. This mimics a fully automated web crawler that would use the actor name as search query and download the top images from IMDB. 
We used one of our unsupervised labelling methods to get an initial labelling, which we corrected ourselves manually to get ground truth labels. On average for each episode, we have 636 tracks with 20584 faces for BBT; 1113 tracks with 20044 faces for BB and 1590 tracks with 29020 faces for MM. These have been labelled as belonging to one of 5/6 main actors in the show, or to a generic side-actor class.
\\
\textbf{Baselines} 
In the next series of experiments we compare our  Hungarian Clustering based self-labelling algorithm referred to as \textbf{HCSL} and its variant without clustering \textbf{HSL} with two direct matching baselines and one self-labelling approach: 

\begin{enumerate}
\item Closest Average Actor ({\bf AVG}):
As explained before, we start from a set of faces for each of the main actors, downloaded from IMDB. The task is to match each track from the TV series to its corresponding actor. So, in the first direct matching baseline, we compute the average distance between the faces in the track and the set of faces that belongs to each actor.  The track is then assigned to its closest actor according to this assignment cost, i.e. the corresponding average distance. If the average distance between the track and its closest actor cloud is bigger than a predefined threshold, the track is considered to belong to a side actor.  
\item Nearest Neighbor Face({\bf 1NN}): this baseline is similar to AVG, but instead of computing the assignment cost of a track to an actor as the average distance between the set of track faces and the set of IMDB actor faces, the cost is simply the minimum distance between a face in the track and its closest face in the set of actor faces.  As with AVG, a track is assigned to the actor with the minimum cost and if that cost is bigger than a threshold it means it is a side actor.
\item Self-labelling with top 10 percentile ({\bf TopTen}):
In order to examine the added value of the Hungarian Matching coupled with the self-labelling strategy, we add another baseline which is a simple self-labelling approach. Instead of selecting the globally optimal tracks to be assigned, we select the tracks such that their assignment cost is in the  top ten percentile of all track costs. We move them to their corresponding actors clouds and repeat the self-labelling process until no track is remaining. To determine the side actors, we examine the cost of assignment for each track at the end of the self-labelling process and if a track has an assignment cost higher than a threshold it is considered a side actor.
\end{enumerate}

Note that we do not compare to other work because our framework is different (we use actor names for weak supervision - see Supplementary Material).

\noindent
\textbf{Choice of graph edge cost}
We compare our proposed normalized edge cost ({\bf NC}) to the Euclidean Distance ({\bf EUC}).
The main difference is that NC favours the selection of tracks that are close to one actor compared to all others, while EUC looks at the absolute distances and thus could select a track that is close to more than one actor. \newline
Accuracy results for all our experiments are in terms of track labels. The track is our measurement unit - all faces in each track have the same label. \newline
\textbf{Parameter Selection}
The threshold for HSL is a tradeoff between assigning hard tracks to IMDB clusters and false assignment of side actor tracks. It is set to the average distance between main actor clusters and side actor tracks in the last iterations of self-labelling. 
All hyperparameters (see Supplementary Material) were selected once (while testing on the first episode of BB) and not changed for the duration of our experiments.

\section{Results  and Discussion}
\label{ResultsAndDiscussion}
\begin{figure}[t]
\vspace*{-10pt}
 \centering
\includegraphics[width=1.2\textwidth]{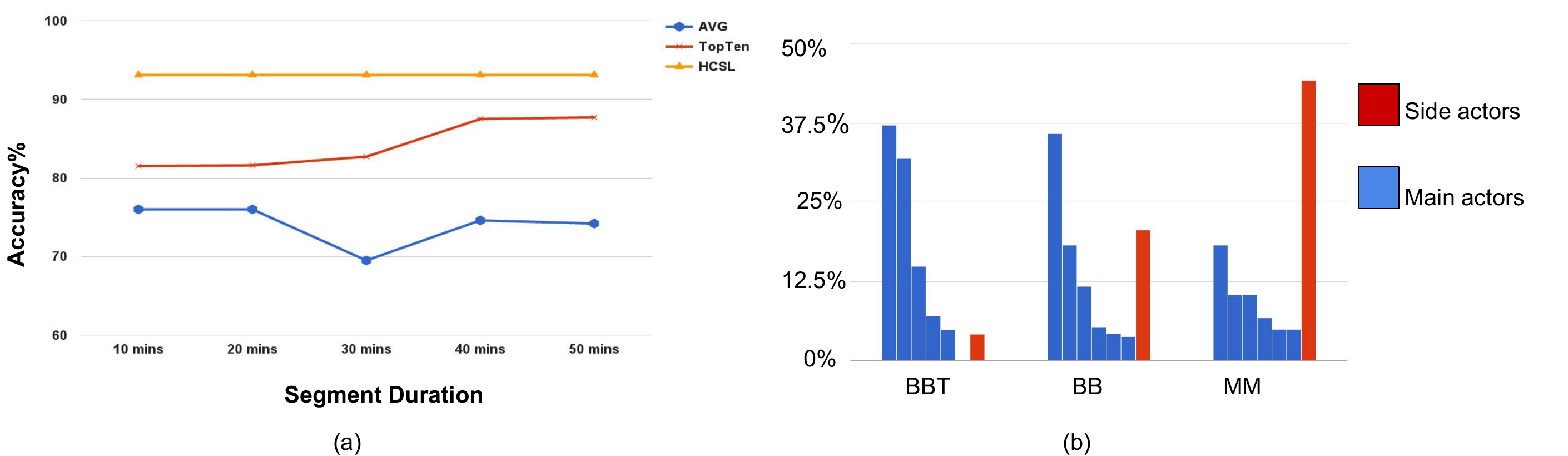}
\caption{(a) Percentage of each main actor and the side actors. (b) Baselines and our method compared on variable length segments in Breaking Bad.}
\label{fig:segments}
\end{figure}

Table \ref{tab1} shows that all methods including baselines achieve above 90\% results averaged over 6 episodes of the \textbf{ Bang Theory (BBT) dataset}. This is because all the main actors have distinctive appearances and also because of the existence of very few side-actors (Figure \ref{fig:segments}(b)). In addition, there is very little change in each actor's appearance over the course of the TV series, except in episode 6, where the actors are wearing masks. Consequently, baseline methods' performance decrease noticeably in episode 6, while our HSL method retains a performance of 90\%. In this case, the clustering HCSL scheme does not seem to improve on top of the Hungarian self-labelling HSL method, probably due to the limited change in actor appearance over the episode and the existence of very few side-actors.


The second TV series, \textbf{Breaking Bad (BB)}, has much more change in actor appearance over the course of the show. For example, the character Walter White develops cancer and shaves his head during chemotherapy. When the character Jesse Pinkman gets hit in the face, he gets a black eye. There is also more variety in the shots in this show, including indoors as well as outdoors, and dimly lit conditions. There is camera shake, leading to blurred faces and noisy tracks. This results in a drop in performance of the direct matching methods (1NN and AVG) from 92\% in  BBT to 77.3\% in BB as reported in Table \ref{tab2}. The Hungarian Self-labelling (HSL) shows a good mean accuracy of 90.3\%. The self labelling with  top ten percentile (TopTen) performs slightly less with an  average of 89.4\%. In contrast to BBT, the Hungarian Clustering based Self-labelling (HCSL) algorithm achieves better results (average accuracy of 93.0\%) in the presence of above mentioned actor appearance change and false detections (outliers).


From Table \ref{tab1} and Table \ref{tab2}, we see that using our normalized measure (the choice of edge cost for each method is shown in column 2), helps improving the performance of the self-labelling methods by 2\%-3\% on average. This illustrates the need for such an edge cost to improve the selection procedure by taking into account the track distance to the other actors.

In the third dataset,\textbf{ Mad Men} TV series, we only report the results of the best direct matching baseline (AVG) and the best self-labelling baseline (TopTen with NC edge cost). 
Mad Men presents a unique situation, where the number of side-actors is greater than the number of main actors (Figure \ref{fig:segments}(b)). In addition, the women in the show are presented both with and without makeup, resulting in appearance changes that made it difficult even for the human annotator to identify them as the same person purely based on facial features. The self-labelling based methods show decreased performance compared to the previous shows, but still improve noticeably on the direct matching baseline that achieves an average of 74.8\%. TopTen gives an average performance of 85.8\%, while our proposed method (HCSL) achieves a performance of 86.1\% on average.\\

\noindent
\textbf{Variable length movies} 
We examine the effect of variable length videos on the performance using segments of increasing length (in 10 minute increments) in BB (Figure \ref{fig:segments}(a)). The Hungarian self-labelling (HCSL) beats the other baselines, and it is important to note that the performance of HCSL remains relatively constant over the variable lengths of videos, in contrast with the TopTen baseline. This is because the selection of tracks in each iteration of the TopTen self-labelling process is dependent on the proportion of tracks belonging to each actor. This effect is more pronounced in segments of shorter duration. The effect of track imbalance between actors is overcome by a larger number of tracks in segments of longer duration. However, HCSL ensures the optimal selection of tracks for every actor in each iteration, regardless of the number of tracks per actor.\\

\begin{table}
 \vspace*{-10pt}
\centering
\begin{tabular}{|c | c | c| c | c | c | c | c |c|}
\hline
Method &Edge Cost&EP1& EP2& EP3 & EP4 & EP5 & EP6 & avg\\  \hline
1NN&Euc&92.0&93.3&95.5&95.4&94.9&83.3&92.4 \\ \hline
AVG&Euc&93.0&92.2&94.1&\bf{97.5}&93.4&82.0&92.0 \\ \hline
Top Ten&Euc &94.8&\bf{97.8}&98.1&91.2&91.9&84.0&92.9 \\ \hline
Top Ten& NC& 94.2 & \bf{97.8} & \bf{99.8 }& 93.7 & 96.6 & 85.8 & 94.6 \\ \hline
HSL(Our)&Euc&\bf{96.0}&97.6&98.8&90.8&96.1&88.1&94.5 \\ \hline
HSL(Our)&NC&95.5&\bf{97.8}&99.7&97.0&\bf{97.1}&\bf{90.0}&\bf{96.1} \\ \hline
HCSL(Our)&NC&95.4&97.4&\bf{99.8}&92.5&96.9&88.4&95.0 \\ \hline

\end{tabular}

\caption{Different methods performance on Big Bang Theory Dataset}

\label{tab1}
\end{table}

\begin{table}
  \vspace{-10pt}
\centering
\begin{tabular}{|c | c | c| c | c | c | c | c |c|}
\hline
Method &Edge Cost&EP1& EP2& EP3 & EP4 & EP5 & EP6 & avg\\  \hline
1NN&Euc&66.9&63.5&58.5&65.4&73.5&70.0&66.3 \\ \hline
AVG&Euc&75.1&74.0&70.9&76.8&82.8&84.5&77.3 \\ \hline
Top Ten &Euc&93.4&90.3&66.9&92.5&90.7&94.5&88.0\\ \hline
Top Ten&NC&95.2&93.8&69.7&92.2&\bf{92.0}&94.0&89.4\\ \hline
HSL(Our)&Euc&94.7&94.3&56.4&\bf{93.4}&91.6&\bf{94.6}&87.5 \\ \hline
HSL(Our)& NC&\bf{95.9}&94.9&72.6&91.9&91.9&\bf{94.6}&90.3\\ \hline
HCSL(Our)&NC&95.5&\bf{95.1}&\bf{88.8}&93.2&91.6&94.3&\bf{93.0} \\ \hline

\end{tabular}
\caption{Baselines and our method compared on Breaking Bad}
  \vspace{-15pt}
\label{tab2}
\end{table}

\begin{figure}
\vspace{-10pt}
 \centering
\includegraphics[width=0.68\textwidth,scale=0.1]{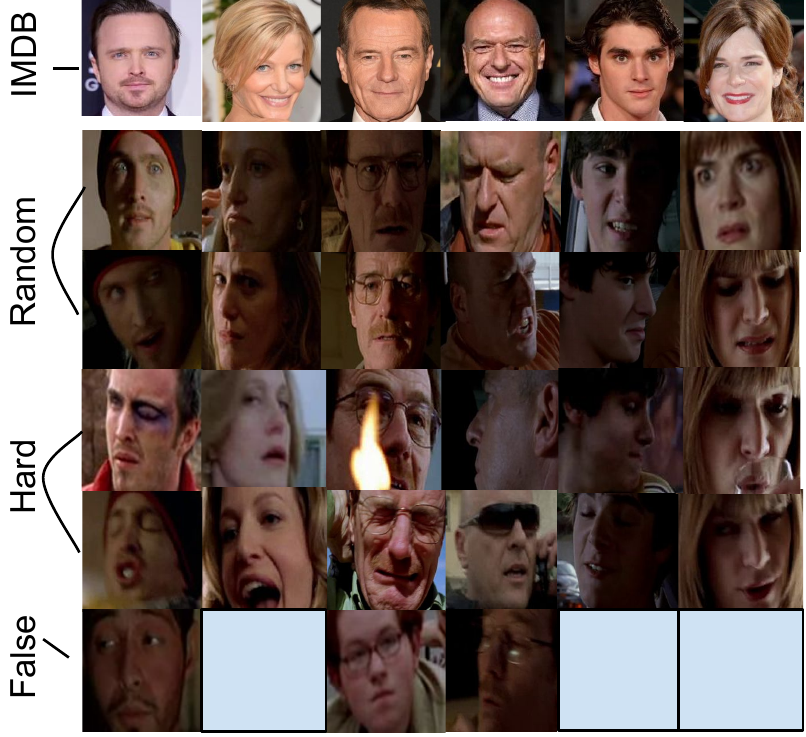}
\caption{Samples of labelled images for the main actors in the Breaking Bad dataset.Top row IMDB images, Rows 2-3: randomly selected images. Rows 4-5: hard cases. Row 6: false labels. Best viewed in pdf.}
\label{fig:LabellingExamples}
  \vspace{-10pt}
\end{figure}

\begin{table}[h]
\centering
\begin{tabular}{ | c | c| c | c | c |}
\hline
Method &Ep1&Ep2&Ep3&avg \\  \hline
AVG&72.4    &72.8&    79.4&    74.8 \\ \hline
Top Ten &\bf{86.2}   & 86.1   & 85.3   & 85.8 \\ \hline
HCSL(our) &85.9   & \textbf{86.6}   & \textbf{86.1}     & \textbf{86.1}\\ \hline 
\end{tabular}
\caption{Baselines and our method compared on Mad Men}
  \vspace{-20pt}
\label{tab3}
\end{table}

\noindent
\textbf{Time Complexity} Here's a breakdown of the computation time of our current system: face detection: 22 seconds (per frame); deep CNN face feature extraction per face: 0.5 seconds (per frame); Tracking: 0.1 second (per frame); HCSL: about the same as the running time of the episode. The complexity is of order $O(n^2)$ where $n$ is the number of tracks. This is dominated by the pairwise distance calculation between the tracks. 
The main bottleneck is the DPM face detection, which could be sped up using GPU.\\

\noindent
\textbf{Qualitative results}
Figure \ref{fig:LabellingExamples} shows sample results of our HCSL algorithm from the Breaking Bad TV series. We present a mixture of randomly selected, hard and wrongly labelled faces per actor. During the initial stages of self-labelling, the algorithm picks the easier faces (more similar to IMDB faces). After a few iterations, the algorithm acquires more faces from the dataset, is better able to bridge the domain shift between IMDB and the TV show, and is able to correctly label the more challenging faces. At the end of the self-labelling process, when the main actor tracks have been exhausted, the algorithm starts assigning the side-actor tracks to the actor clouds. During this stage of the self-labelling process, the edge cost threshold is the only thing controlling the assignment process. This threshold (as mentioned earlier) is fixed for all our experiments, and might not be the optimal one for the particular show/actor combination.

\section{Conclusions and Future Work}
\label{Conclusion}
In this work we utilize the power of deep CNN features to automatically label actors in TV series, using their photos from IMDB, starting only from the names of the lead actors as weak supervision.
There is a domain shift between actor images in IMDB and their appearance in TV series. We overcome this problem by introducing a graph-matching based self-labelling approach that iteratively adds actor faces from the TV series to the original collection of actor photos from IMDB. We obtain an appearance profile for each actor based on clustering the self-labelled actor images. 
We believe that our method is generic enough to be applied to 
TV news broadcasts, documentary films, silent movies and home videos, given only a few representative images of the actors therein. In future work, we plan on including side actors in the graph matching procedure by automatically building side actor templates which should further decrease the number of false positives. 
We will also explore metric learning to learn a good similarity measure from the data.
\newline 
\footnotesize {\textbf{Acknowledgment:}
This work was supported by the iMinds HiViz project. The first author's PhD is funded by an FWO scholarship.}




\bibliographystyle{splncs}
\bibliography{egbib}

\begin{thebibliography}{10}

\bibitem{everingham2006hello}
Everingham, M., Sivic, J., Zisserman, A.:
\newblock Hello! my name is... buffy''--automatic naming of characters in tv
  video.
\newblock In: BMVC. Volume~2. (2006) ~6

\bibitem{tapaswi2012knock}
Tapaswi, M., B{\"a}uml, M., Stiefelhagen, R.:
\newblock “knock! knock! who is it?” probabilistic person identification in
  tv-series.
\newblock In: Computer Vision and Pattern Recognition (CVPR), 2012 IEEE
  Conference on, IEEE (2012)  2658--2665

\bibitem{hu2015deep}
Hu, Y., Ren, J.S., Dai, J., Yuan, C., Xu, L., Wang, W.:
\newblock Deep multimodal speaker naming.
\newblock In: Proceedings of the 23rd Annual ACM Conference on Multimedia
  Conference, ACM (2015)  1107--1110

\bibitem{ren2016look}
Ren, J., Hu, Y., Tai, Y.W., Wang, C., Xu, L., Sun, W., Yan, Q.:
\newblock Look, listen and learn-a multimodal lstm for speaker identification.
\newblock arXiv preprint arXiv:1602.04364 (2016)

\bibitem{Taigman2014deepface}
Taigman, Y., Yang, M., Ranzato, M., Wolf, L.:
\newblock Deepface: Closing the gap to human-level performance in face
  verification.
\newblock In: Proceedings of the IEEE Conference on Computer Vision and Pattern
  Recognition. (2014)  1701--1708

\bibitem{Parkhi15}
Parkhi, O.M., Vedaldi, A., Zisserman, A.:
\newblock Deep face recognition.
\newblock In: British Machine Vision Conference. (2015)

\bibitem{schroff2015facenet}
Schroff, F., Kalenichenko, D., Philbin, J.:
\newblock Facenet: A unified embedding for face recognition and clustering.
\newblock In: Proceedings of the IEEE Conference on Computer Vision and Pattern
  Recognition. (2015)  815--823

\bibitem{sivic2009you}
Sivic, J., Everingham, M., Zisserman, A.:
\newblock “who are you?”-learning person specific classifiers from video.
\newblock In: Computer Vision and Pattern Recognition, 2009. CVPR 2009. IEEE
  Conference on, IEEE (2009)  1145--1152

\bibitem{bauml2013semi}
Bauml, M., Tapaswi, M., Stiefelhagen, R.:
\newblock Semi-supervised learning with constraints for person identification
  in multimedia data.
\newblock In: Proceedings of the IEEE Conference on Computer Vision and Pattern
  Recognition. (2013)  3602--3609

\bibitem{tapaswi2015improved}
Tapaswi, M., Bauml, M., Stiefelhagen, R.:
\newblock Improved weak labels using contextual cues for person identification
  in videos.
\newblock In: Automatic Face and Gesture Recognition (FG), 2015 11th IEEE
  International Conference and Workshops on. Volume~1., IEEE (2015)  1--8

\bibitem{bojanowski2013finding}
Bojanowski, P., Bach, F., Laptev, I., Ponce, J., Schmid, C., Sivic, J.:
\newblock Finding actors and actions in movies.
\newblock In: Proceedings of the IEEE International Conference on Computer
  Vision. (2013)  2280--2287

\bibitem{Parkhi15a}
Parkhi, O.M., Rahtu, E., Zisserman, A.:
\newblock It's in the bag: Stronger supervision for automated face labelling.
\newblock In: ICCV Workshop: Describing and Understanding Video \& The Large
  Scale Movie Description Challenge, IEEE (2015)

\bibitem{hauriletnaming}
Haurilet, M.L., Tapaswi, M., Al-Halah, Z., Stiefelhagen, R.:
\newblock {Naming TV Characters by Watching and Analyzing Dialogs}.
\newblock In: IEEE Winter Conference on Applications of Computer Vision (WACV).
  (2016)

\bibitem{guillaumin2010multiple}
Guillaumin, M., Verbeek, J., Schmid, C.:
\newblock Multiple instance metric learning from automatically labeled bags of
  faces.
\newblock In: Computer Vision--ECCV 2010.
\newblock Springer (2010)  634--647

\bibitem{zhu2002learning}
Zhu, X., Ghahramani, Z.:
\newblock Learning from labeled and unlabeled data with label propagation.
\newblock Technical report, Citeseer (2002)

\bibitem{tuytelaars2011naming}
Pham, Phi~T., T.T., Moens, M.F.:
\newblock Naming people in news videos with label propagation.
\newblock IEEE multimedia \textbf{18} (2011)  44--55

\bibitem{kumar2014face}
Kumar, V., Namboodiri, A.M., Jawahar, C.:
\newblock Face recognition in videos by label propagation.
\newblock In: 2014 22nd International Conference on Pattern Recognition (ICPR),
  IEEE (2014)  303--308

\bibitem{patel2015visual}
Patel, V.M., Gopalan, R., Li, R., Chellappa, R.:
\newblock Visual domain adaptation: A survey of recent advances.
\newblock Signal Processing Magazine, IEEE \textbf{32} (2015)  53--69

\bibitem{bruzzone2010domain}
Bruzzone, L., Marconcini, M.:
\newblock Domain adaptation problems: A dasvm classification technique and a
  circular validation strategy.
\newblock Pattern Analysis and Machine Intelligence, IEEE Transactions on
  \textbf{32} (2010)  770--787

\bibitem{habrard2013iterative}
Habrard, A., Peyrache, J.P., Sebban, M.:
\newblock Iterative self-labeling domain adaptation for linear structured image
  classification.
\newblock International Journal on Artificial Intelligence Tools \textbf{22}
  (2013)  1360005

\bibitem{banerjee2015novel}
Banerjee, B., Bovolo, F., Bhattacharya, A., Bruzzone, L., Chaudhuri, S.,
  Buddhiraju, K.M.:
\newblock A novel graph-matching-based approach for domain adaptation in
  classification of remote sensing image pair.
\newblock Geoscience and Remote Sensing, IEEE Transactions on \textbf{53}
  (2015)  4045--4062

\bibitem{tuia2013graph}
Tuia, D., Mu{\~n}oz-Mar{\'\i}, J., G{\'o}mez-Chova, L., Malo, J.:
\newblock Graph matching for adaptation in remote sensing.
\newblock Geoscience and Remote Sensing, IEEE Transactions on \textbf{51}
  (2013)  329--341

\bibitem{mathias2014face}
Mathias, M., Benenson, R., Pedersoli, M., Van~Gool, L.:
\newblock Face detection without bells and whistles.
\newblock In: Computer Vision--ECCV 2014.
\newblock Springer (2014)  720--735

\bibitem{voc-release5}
Girshick, R.B., Felzenszwalb, P.F., McAllester, D.:
\newblock Discriminatively trained deformable part models, release 5.
\newblock (http://people.cs.uchicago.edu/~rbg/latent-release5/)

\bibitem{kuhn1955hungarian}
Kuhn, H.W.:
\newblock The hungarian method for the assignment problem.
\newblock Naval research logistics quarterly \textbf{2} (1955)  83--97

\bibitem{Long_2014_CVPR}
Long, M., Wang, J., Ding, G., Sun, J., Yu, P.S.:
\newblock Transfer joint matching for unsupervised domain adaptation.
\newblock In: The IEEE Conference on Computer Vision and Pattern Recognition
  (CVPR). (2014)

\bibitem{ward1963hierarchical}
Ward~Jr, J.H.:
\newblock Hierarchical grouping to optimize an objective function.
\newblock Journal of the American statistical association \textbf{58} (1963)
  236--244

\bibitem{bubeck2009nearest}
Bubeck, S., Luxburg, U.v.:
\newblock Nearest neighbor clustering: A baseline method for consistent
  clustering with arbitrary objective functions.
\newblock The Journal of Machine Learning Research \textbf{10} (2009)  657--698

\end{thebibliography}

\end{document}